\documentclass{article}

\usepackage[preprint,nonatbib]{neurips_2020}

\usepackage[utf8]{inputenc} 
\usepackage[T1]{fontenc}    
\usepackage{hyperref}       
\usepackage{url}            
\usepackage{booktabs}       
\usepackage{amsfonts}       
\usepackage{nicefrac}       
\usepackage{microtype}      

\usepackage{threeparttable,color}
\usepackage{booktabs}
\usepackage{multirow}
\usepackage{enumitem}
\usepackage{subfig}

\usepackage{wrapfig}
\usepackage{subfig}
\usepackage{enumitem}
\usepackage{hyperref}
\hypersetup{
    colorlinks=true,
    linkcolor=black,
    filecolor=black,      
    urlcolor=blue,
    citecolor=black,
}
\usepackage{cite}
\usepackage{amsmath,amssymb,amsfonts}
\usepackage{algorithmic}
\usepackage{graphicx}
\usepackage{textcomp}
\usepackage{xcolor}
\def\BibTeX{{\rm B\kern-.05em{\sc i\kern-.025em b}\kern-.08em
    T\kern-.1667em\lower.7ex\hbox{E}\kern-.125emX}}

\title{Self-supervised Learning on Graphs: \\ Deep Insights and New Directions}

\author{%
    Wei Jin \\
    Michigan State University\\
    \texttt{jinwei2@msu.edu} \\
   \And
    Tyler Derr \\
    Michigan State University\\
    \texttt{derrtyle@msu.edu} \\
  \And
    Haochen Liu \\
    Michigan State University\\
    \texttt{liuhaoc1@msu.edu} \\
   \And
    Yiqi Wang \\
    Michigan State University\\
    \texttt{wangy206@msu.edu} \\ 
   \And
    Suhang Wang \\
    The Pennsylvania State University\\
    \texttt{szw494@psu.edu} \\ 
    \And
    Zitao Liu \\
    TAL Education Group \\
    \texttt{liuzitao@100tal.com} \\ 
   \And
       Jiliang Tang \\
    Michigan State University\\
    \texttt{tangjili@msu.edu} \\ 
}

\begin{document}

\maketitle

\begin{abstract}

The success of deep learning notoriously requires larger amounts of costly annotated data. This has led to the development of self-supervised learning (SSL) that aims to alleviate this limitation by creating domain specific pretext tasks on unlabeled data. Simultaneously, there are increasing interests in generalizing deep learning to the graph domain in the form of graph neural networks (GNNs). GNNs can naturally utilize unlabeled nodes through the simple neighborhood aggregation that is unable to thoroughly make use of unlabeled nodes. Thus, we seek to harness SSL for GNNs to fully exploit the unlabeled data. Different from data instances in the image and text domains, nodes in graphs present unique structure information and they are inherently linked indicating not independent and identically distributed (or i.i.d.). Such complexity is a double-edged sword for SSL on graphs. On the one hand, it determines that it is challenging to adopt solutions from the image and text domains to graphs and dedicated efforts are desired. On the other hand, it provides rich information that enables us to build SSL from a variety of perspectives. Thus, in this paper, we first deepen our understandings on when, why, and which strategies of SSL work with GNNs by empirically studying numerous basic SSL pretext tasks on graphs. Inspired by deep insights from the empirical studies, we propose a new direction {\it SelfTask} to build advanced pretext tasks that are able to achieve state-of-the-art performance on various real-world datasets. The specific experimental settings to reproduce our results can be found in \url{https://github.com/ChandlerBang/SelfTask-GNN}.
\end{abstract}

\section{Introduction}
In recent years, deep learning has achieved superior performance across numerous domains; but it requires costly annotations of huge amounts of data~\cite{kolesnikov2019revisiting}. Hence, self-supervised learning (SSL) has been introduced in both the image~\cite{kolesnikov2019revisiting,doersch2015unsupervised, caron2018deepcluster} and text~\cite{le2014word2vec,devlin2018bert} domains to alleviate the need of large labeled data by deriving labels for the significantly more unlabeled data. More specifically, SSL often first designs a domain specific pretext task to assign labels for data instances and then trains the deep model on the pretext task to learn better representations due to the inclusion of unlabeled samples in the training process.

As the generalization of deep learning to the graph domain, graph neural networks (GNNs) have been proven to be powerful in graph representation learning. As a result, GNNs have facilitated various computational tasks on graphs such as node classification and graph classification~\cite{wu2019comprehensive-survey,kipf2016semi,hamilton2017inductive,gat}. In this work, we focus on advancing GNNs for node classification where GNNs leverage both labeled and unlabeled nodes on a graph to jointly learn node representations and a classifier that can predict the labels of unlabeled nodes on the graph. On the one hand, GNNs are inherently semi-supervised where unlabeled data has been coherently integrated. On the other hand, GNNs mainly utilize unlabeled nodes by simply aggregating their features that cannot thoroughly take advantage of the abundant information~\cite{sun2019m3s}. Thus, to fully exploit the unlabeled nodes for GNNs, SSL can be naturally harnessed for providing additional supervision.

Graph-structured data is often more complex than other domains (e.g., image and text). In addition to node attributes, graphs present complicated structure information. For example, the topology of an image is a fixed grid and text is a simple sequence, while graphs are not restricted to these rigid structures. Furthermore, unlike images and text where the entire structure is a single data sample, each node in a graph is an individual instance and has its own associated attributes and topological structures. The complexity of graph-structured data does not stop here. In the text and image domain, data samples are often under the assumption of being {\it i.i.d.} (independent and identically distributed). However, in the graph domain, instances (or nodes) are inherently linked and dependent of each other. Therefore, the complex nature of graph-structured data determines that it is very challenging to directly adopt self-supervised learning developed in other domains to graphs. While introducing tremendous challenges, the complexity of graphs is a double-edged sword that also presents unprecedented opportunities. In particular, the complexity provides rich information that enables us to design pretext tasks from various perspectives. Similar to the image and text domains, we can focus on individual nodes such as node features and node topological properties. Moreover, unlike the image and text domains, nodes are dependent in a graph, and thus we are able to investigate new aspects such as dependence on node pairs or even a set of nodes. In addition, multiple information resources including node attributes, structure information, and label information of labeled nodes are available in a graph and their interactions and combinations provide unprecedented opportunities for us to design advanced self-supervised pretext tasks. Very recently, there are only a few attempts to adapt SSL from the image domain in training graph neural networks~\cite{sun2019m3s,peng2020self}. Therefore, the research of self-supervised learning on graphs is still at the initial stage and more systematical and dedicated efforts are pressingly needed.

In this paper, we embrace the challenges and opportunities to study self-supervised learning in graph neural networks for node classification with two major goals.  First, we want to deepen our understandings on self-supervised learning on graphs. Specifically, there are a variety of potential pretext tasks for graphs; hence it is important to gain insights on {\it when} and {\it why} SSL works for GNNs and {\it which} strategy can better integrate SSL for GNNs. Second, we target on inspiring new directions of SSL on graphs according to our understandings. Particularly, we want to investigate {\it how} these insights can motivate more sophisticated approaches to design pretext tasks.  To achieve the first goal, we design basic types of pretext tasks directly based on attribute and structural information. We make several crucial findings about SSL on graphs via deep analysis on their impact on the GNN performance. These findings allow us to propose a new direction \textit{SelfTask} to design more advanced pretext tasks that are empirically demonstrated to achieve state-of-the-art performance on various graph datasets.

\section{Problem Statement}\label{sec:problem}
We use $\mathcal{G}=({\bf\mathcal{V}},{\bf\mathcal{E}}, {\bf X})$ to denote a graph where $\mathcal{V}=\{v_1, v_2, ..., v_N\}$ is the set of $N$ nodes, $\mathcal{E}$ is the set of edges describing the relations between nodes and ${\bf X} = [{\bf x}_1,{\bf x}_2,\ldots,{\bf x}_N]$ is the node feature matrix where ${\bf x}_i$ denotes the node features of $v_i$. The graph structure information can also be represented by an adjacency matrix $\mathbf{A} \in [0,1]^{N \times N}$ where $\mathbf{A}_{ij}=1$ indicates there exists a link between nodes $v_i$ and $v_j$, otherwise $\mathbf{A}_{ij}=0$. Hence, a graph can also be denoted as $\mathcal{G}=({\bf A},{\bf X})$. 
In this paper, we focus on the semi-supervised node classification setting where only a subset of nodes $\mathcal{V}_{L}$ are associated with corresponding labels $\mathcal{Y}_L$. We denote the labeled data as $\mathcal{D}_L=(\mathcal{V}_{L},\mathcal{Y}_L)$ and unlabeled data as $\mathcal{D}_{U}$. Let $f_\theta: \mathcal{V}_L\rightarrow\mathcal{Y}_L$ be a
graph neural network that maps the nodes to the set of labels such that the graph neural network can infer the labels of unlabeled data. Thus, the objective function for the semi-supervised node classification task can be formulated as minimizing the loss $\mathcal{L}_{task}$, or more specifically as
\begin{equation}
\min _{\theta} \mathcal{L}_{task}( {\theta}, {\bf A}, {\bf X}, \mathcal{D}_L)=\sum_{(v_i, y_i) \in \mathcal{D}_{L}} \ell\left(f_{\theta}(\mathcal{G})_{v_i}, y_{i}\right),
\end{equation}
where $\theta$ is used to denote the parameters of $f_\theta$, $f_\theta( \mathcal{G})_{v_i}$ is the prediction of node $v_i$ and $\ell(\cdot, \cdot)$ denotes the loss function used to measure the difference between the predicted and true labels (e.g., cross entropy).  

With the aforementioned notations and definitions, there could be multiple settings of SSL for GNN, but in this work we formally define the problem of self-supervised learning for graph neural networks under the task of node classification as: 
\vskip 0.5em

\textbf{Problem 1.} \textit{Given a dataset in the graph domain represented as a graph $\mathcal{G}=({\bf A, X})$ with paired labeled data $\mathcal{D}_L=(\mathcal{V}_{L},\mathcal{Y}_L)$, we aim to construct a self-supervised pretext task with a corresponding loss $\mathcal{L}_{self}$ that can be integrated with the task specific loss $\mathcal{L}_{task}$ to learn a graph neural network $f_\theta$ that can better generalize on the unlabeled data.}

\section{Basic Pretext Tasks on Graphs}\label{sec:sslinfo}

In this section, we present various types of self-supervised pretext tasks on graphs. More specifically, we investigate defining pretext tasks based upon self-supervised information from: (A) the underlying graph structure information (i.e., $\mathbf{A}$); or (B) node feature/attribute information (i.e., $\mathbf{X}$). These two directions are the most natural sources of information to develop self-supervised information for the unlabeled nodes. As there can be a variety of potential pretext tasks for graphs, we first provide detailed justifications for each of the self-supervised pretext tasks and present the details of representative methods in both (A) and (B).  Thereafter, in Section~\ref{sec:ssladvanced} we present more advanced pretext tasks built upon deep insights from an empirical study in Section~\ref{sec:sslstudy}.

\subsection{Structure Information}
The first natural choice for extracting self-supervised information in the graph domain is the inherent structure behind the data. This is because unlike the image and text domains, in graphs our data instances are related (i.e., the nodes are linked together). Thus, one main direction is to construct self-supervision information for the unlabeled nodes based on their local structure information, or how they relate to the rest of the graph. In other words, the structure information for establishing self-supervised pretext tasks can be categorized into either local or global structure information. 

\subsubsection{Local Structure Information} 
From the local perspective of developing self-supervised information, it can either come from the node itself, or from the structural relationship that node has in its local surrounding neighborhood. In addition, the pretext task can be defined on a single node, or can be developed in a pairwise/contrastive way that involves combining/comparing the information from more than one node. Next, we present two representative examples of local structure based SSL pretext tasks considering these different aspects.

\begin{itemize}[leftmargin=*]
        \item \textbf{NodeProperty}. In this task, we aim to predict the property for each node in the graph such as their degree, local node importance, and local clustering coefficient. The goal of this pretext task is to (further) encourage the GNN to learn local structure information in addition to the specific task that is being optimized. In this work, we use node degree as a representative local node property for self-supervision while leaving other node properties (or the combination) as one future work. More formally, we let $d_i = \sum_{j=1}^N{\mathbf{A}_{ij}}$ denote the degree of $v_i$ and construct the associated loss of the self-supervised pretext task as  
    \begin{align}
        \mathcal{L}_{self}( {\theta}^{\prime}, {\bf A}, {\bf X}, \mathcal{D}_U)= \frac{1}{\vert\mathcal{D}_{U}\vert}\sum_{v_i \in \mathcal{D}_{U}} (f_{\theta^{\prime}}(\mathcal{G})_{v_i} - d_{i})^2,
    \end{align}
    where $\theta^{\prime}$ is used to denote the parameters of a graph neural network model $f_{\theta^{\prime}}$, $\mathcal{D}_{U}$ represents the set of unlabeled nodes and associated pretext task labels in the graph, and $f_{\theta^{\prime}}(\mathcal{G})_{v_i}$ is used to denote the the predicted local node property for node $v_i$ (which in this case is the predicted node degree). The intuition of constructing self-supervised pretext tasks related to the local node property is to ultimately guide the features (i.e., node representations) from the GNN to preserve this information. This relies on the assumption that such node property information is related to the specific task of interest.
    \item \textbf{EdgeMask.}     
    For the edge mask task, we seek to develop the self-supervision based not only on an individual node itself, but instead pairwise, based on the connections between two nodes in the graph. In particular, we first randomly mask some edges and then the model is asked to reconstruct the masked edges. More specifically, we first mask $m_e$ edges denoted as the set $\mathcal{M}_e  \subset \mathcal{E}$ and also sample the set $\overline{\mathcal{M}}_e = \{(v_i,v_j)| v_i, v_j \in \mathcal{V} \text{ and } (v_i, v_j) \notin \mathcal{E}\}$ of node pairs of equal size (i.e., $|\overline{\mathcal{M}}_e| = |\mathcal{M}_e| = m_e$). Then, the SSL pretext task here is to predict whether or not there exists a link between a given node pair. More formally, we construct the associated loss as
    \begin{align}
    \mathcal{L}_{self}&( {\theta}^{\prime}, {\bf A}, {\bf X}, \mathcal{D}_U) = \\ \nonumber
    & \frac{1}{\vert\mathcal{M}_e\vert}\sum_{(v_i,v_j) \in \mathcal{M}_e} \ell\left( f_{w}(\vert f_{\theta^{\prime}}(\mathcal{G})_{v_i} - f_{\theta^{\prime}}(\mathcal{G})_{v_j}\vert), 1\right) \\ \nonumber
    + & \frac{1}{\vert\overline{\mathcal{M}}_e\vert}\sum_{(v_i,v_j) \in \overline{\mathcal{M}}_e} \ell\left( f_{w}(\vert f_{\theta^{\prime}}(\mathcal{G})_{v_i} - f_{\theta^{\prime}}(\mathcal{G})_{v_j}\vert), 0\right),
    \end{align}
    where $f_{\theta^{\prime}}(\mathcal{G})_{v_i}$ denotes the embedding of node $v_i$,  $\ell(\cdot, \cdot)$ is the cross entropy loss, $f_{w}$ linearly maps to 1-dimension, and the class of having a link between $v_i$ and $v_j$ is indicated by 1 and 0 otherwise. In summary, this method is expected to help GNN learn information about local connectivity. 
\end{itemize}
    
 \subsubsection{Global Structure Information}
Global self-supervision information for a given node is not only based on the node itself or limited to its immediate local neighborhood, but also taking a bird's-eye view of the position of the node in the graph. Similar to the local perspective, we also propose two representative SSL pretext tasks where one is based upon a global pairwise comparison between two nodes and the other is from how a single node is globally positioned in the graph.
\begin{itemize}[leftmargin=*]
    \item \textbf{PairwiseDistance.} The {\it EdgeMask} pretext task is from a local structure perspective based on masking and trying to recover/predict local edges in the graph. We further develop the {\it PairwiseDistance} where we aim to guide the graph neural network to maintain global topology information through a pairwise comparison. In other words, the pretext task is designed to be able to distinguish/predict the distance between different node pairs. We note that distance can be measured in a variety of ways, such as being in the same connected component/cluster or not, personalized PageRank or other global link prediction methods that calculate node similarity~\cite{liben2007link}, etc. In this work, similar to global context prediction in~\cite{peng2020self}, we elect to use the shortest path length as a measure of the distance between nodes. More specifically, we first calculate the pairwise node shortest path length $p_{ij}$ for all node pairs $\{(v_i,v_j)| v_i, v_j \in \mathcal{V}\}$ and further group the lengths into four categories -- $p_{ij}=1, p_{ij}=2, p_{ij}=3$, and $p_{ij}\geq4$. The reasons of selecting four bins for the path length between two nodes is that the GNN should be able to correctly judge the distance between two nodes to some extent, but if we were to include more classes it would: 1) require more calculations to discover all the actual pairwise distances (if greater than 4); and 2) potentially overfit to some of the longer pairwise distances in the graph, which become quite noisy as compared to the shorter path lengths. In addition, since using all node pairs in the objective would be computationally expensive during the training process, in practice, we randomly sample a certain amount of node pairs $\mathcal{S}$ 
    used for self-supervision during each epoch. The SSL loss can then be formulated as a multi-class classification problem as follows,
    \begin{align}
    \mathcal{L}_{self}&( {\theta}^{\prime}, {\bf A}, {\bf X}, \mathcal{D}_U) = \\ \nonumber
    & \frac{1}{\vert\mathcal{S}\vert}\sum_{(v_i,v_j) \in \mathcal{S}} \ell\left( f_{w}(\vert f_{\theta^{\prime}}(\mathcal{G})_{v_i} - f_{\theta^{\prime}}(\mathcal{G})_{v_j}\vert), C_{p_{ij}} \right) \nonumber
    \end{align}   
    where $C_{p_{ij}}$ is the corresponding distance category of $p_{ij}$, $\ell(\cdot,\cdot)$ denotes the  cross entropy loss, and $f_{w}$ linearly maps to 1-dimension,. Note that we leave other pairwise distance measures and other settings for the shortest path distance as future work.

    \item \textbf{Distance2Clusters.} Although \textit{PairwiseDistance} applies a sampling strategy to reduce time complexity, it is still very time-consuming since we need to calculate pairwise distance for all node pairs. Instead, we derive a new SSL pretext task exploring global structure information by predicting the distance (again in terms of shortest path length) from the unlabeled nodes to predefined graph clusters. This will force the representations to learn a global positioning vector of each of the nodes. In other words, rather than a node predicting the distance pairwise to an arbitrary other node in the graph, instead, we establish a fixed set of anchor/center nodes associated with graph clusters and then each node will predict its distance to this set of anchor nodes. Concretely, we first partitioning the graph to get $k$ clusters $\{C_1,C_2,...,C_k\}$ by applying the METIS graph partitioning algorithm~\cite{karypis1998fast-metis}, since it is commonly used in the literature. Inside each cluster $C_j$, we assign the node with the highest degree to be the center of the corresponding cluster, denoted as $c_{j}$. Then, we can efficiently create a cluster distance vector ${\bf d}_i\in{\mathbb{R}^{k}}$ for node $v_i$ where the $j$-th element of ${\bf d}_i$ is the distance from $v_i$ to the center of $C_j$. The SSL goal of {\it Distance2Clusters} is thus to predict this distance vector and the optimization problem can be formulated as a multiple regression problem as,
    \begin{equation}
    \mathcal{L}_{self}( {\theta}^{\prime}, {\bf A}, {\bf X}, \mathcal{D}_U) = \frac{1}{\vert{\mathcal{D}_U}\vert}\sum_{ v_i \in \mathcal{D}_U} \|f_{\theta^{\prime}}(\mathcal{G})_{v_i} - {\bf d}_{i}\|^2. 
    \end{equation}   
\end{itemize}
    
\subsection{Attribute Information}
In this subsection, we focus on attribute information as the second natural choice for establishing a self-supervised pretext task. Here the key point behind attribute information is to help guide the GNN in a way to ensure certain aspects of node/neighborhood attribute information is encoded in the node embeddings after a self-supervised attribute-based pretext. Next, we design two attribute based pretext tasks. 
\begin{itemize}[leftmargin=*]
    \item \textbf{AttributeMask.} This task is similar to \textit{EdgeMask} but we hope GNN can learn more attribute information via SSL. 
    Thus, we randomly mask (i.e., set equal to zero) the features of $m_a$ nodes $\mathcal{M}_a \subset \mathcal{V}$ where $\vert\mathcal{M}_a\vert$=$m_a$, and then ask the self-supervised component to reconstruct these features. More formally, 
     \begin{align}
    \mathcal{L}_{self}&( {\theta}^{\prime}, {\bf A}, {\bf X}, \mathcal{D}_U) = \frac{1}{\vert{\mathcal{M}_a}\vert}\sum_{ v_i \in \mathcal{M}_a} \|f_{\theta^{\prime}}(\mathcal{G})_{v_i} - {\bf x}_{i}\|^2 \nonumber
    \end{align}   
    However, the features in most real-world datasets are often high-dimensional and sparse. Hence, in practice we first employ Principle Component Analysis (PCA) to obtain reduced dense features before applying AttributeMask. 
    
    \item \textbf{PairwiseAttrSim.} As compared to data samples in other domains such as an image, in graph structured data the aggregation process is actually merging the features from multiple instances to discover the learned representation. Thus, given two nodes that have similar attributes, their learned representations are not necessarily similar (as compared to e.g., two exact images will obtain the same representation in typical deep learning models). More generally, the similarity two nodes have in the input feature space is not guaranteed in the learned representations due to the GNN aggregating features from the two nodes local neighborhoods. This can create a double-edged sword as although we wish to utilize the local neighborhood in a GNN to enhance the node feature transformation, we still wish to somewhat maintain the notion of data instance similarity and not allow a node's neighborhood to drastically change their attribute signature. Thus, we establish the attribute-based SSL pretext task of node attribute similarity.  Due to the majority of the pairwise similarity being near zero, we develop the following pair sampling strategy. First, we let $\mathcal{T}_s$ and $\mathcal{T}_d$ denote the sets of node pairs having the highest similarity and dissimilarity, respectively, which we more formally define as,
    
    \vspace{-2ex}
    \begin{small}
    \begin{align}
    &\mathcal{T}_s = \{(v_i, v_j)| s_{ij} \text{ in top-K of } \{s_{ik}\}^N_{k=1}\setminus\{s_{ii}\},  \forall v_i \in \mathcal{V}_U \} \nonumber \\
    &\mathcal{T}_d = \{(v_i, v_j)| s_{ij} \text{ in bottom-K of } \{s_{ik}\}^N_{k=1}\setminus\{s_{ii}\},  \forall v_i \in \mathcal{V}_U \} \nonumber
    \end{align}
    \end{small}
    \vspace{-2ex}
    
    where $s_{ij}$ measures the node feature similarity between $v_i$ and $v_j$ (according to cosine similarity) and $K$ is the number of top/bottom pairs selected for each node. Now, we can formalize the regression problem as follows,
    \begin{align}
    \mathcal{L}_{self}&( {\theta}^{\prime}, {\bf A}, {\bf X}, \mathcal{D}_U) = \\ \nonumber
    & \frac{1}{\vert\mathcal{T}\vert}\sum_{(v_i,v_j) \in \mathcal{T}} \| f_{w} (\vert f_{\theta^{\prime}}(\mathcal{G})_{v_i} - f_{\theta^{\prime}}(\mathcal{G})_{v_j}\vert) - {s_{ij}} \|^2. \nonumber
    \end{align}
    where $\mathcal{T}=\mathcal{T}_s \cup \mathcal{T}_d$ and $f_{w}$ linearly maps to 1-dimension.
\end{itemize}

\section{Preliminary Analysis}\label{sec:sslstudy}

In the last section, we discussed basic self-supervised pretext tasks from both structure and attribute information. In this section, we present two strategies to merge these pretext tasks into GNNs, i.e., joint training and two-stage training, and then empirically analyze the impacts of the pretext tasks on GNNs.  

\begin{figure}[t]
    \centering
    \includegraphics[width=0.9\linewidth]{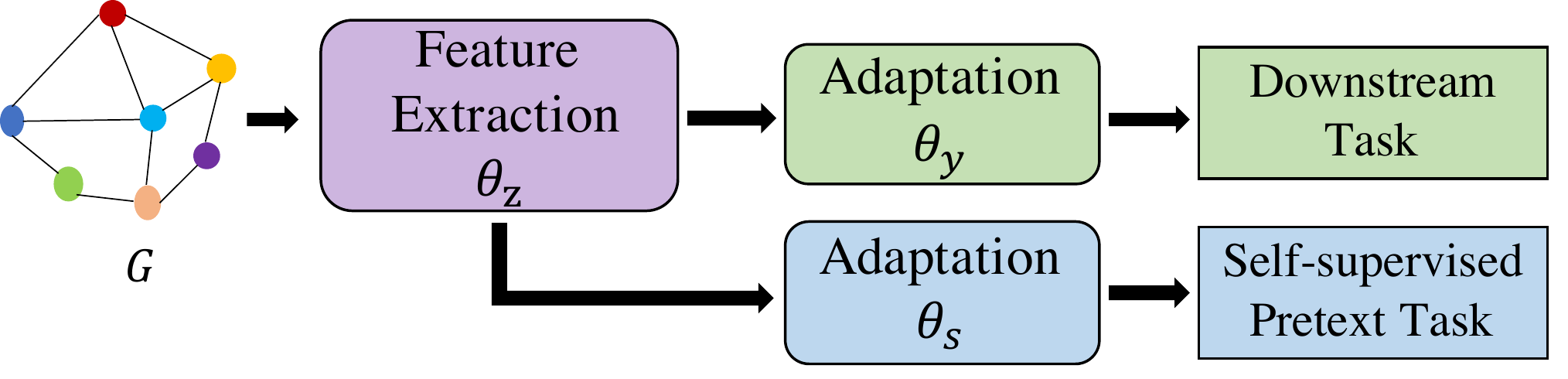}
    \caption{An overview of SSL in GNNs using joint training.}
    \label{fig:joint_training}
\end{figure}

\subsection{Joint Training}
A natural idea to employ self-supervised learning for graph neural networks is to jointly train the corresponding losses. In other words, we aim to optimize the self-supervised loss (i.e., $\mathcal{L}_{self}$) and supervised loss (i.e., $\mathcal{L}_{task}$) simultaneously. 

An overview of joint training is shown in Figure~\ref{fig:joint_training}. Essentially this can be separated into two stages: 1) feature extraction process; and 2) adaptation processes for both the downstream and self-supervised pretext tasks. The first step is a feature extraction process that is applied on the input graph, which can be various graph convolutional layers. Based on the extracted features, two adaptation processes are applied to the downstream task and self-supervised pretext task, respectively. Note that the adaptation layers can be graph convolutional layers or linear layers (which we later discuss in the experiment setup). Afterwards we jointly optimize the losses from both self-supervised and downstream task components.

As introduced in Section~\ref{sec:problem}, we denoted the prediction of a node $v_i$ as $f_\theta( \mathcal{G})_{v_i}$, where $f_\theta$ represented our graph neural network model. As demonstrated in Figure~\ref{fig:joint_training}, we separate the GNN into both a feature extractor and an adapter for the downstream classification task. Correspondingly, we split the parameter $\theta$ as $\theta =\{\theta_z,\theta_y\}$. Then, we
use $f_{\theta_z}(\mathcal{G}) \rightarrow \mathbf{Z}$ to  denote the feature extractor component of $f_\theta$ where $\mathbf{Z}\in \mathbb{R}^{N \times d}$ and $\mathbf{z}_i=f_{\theta_z}(\mathcal{G})_{v_i}$ represents the embedding of node $v_i$. Furthermore, we utilize $f_{\theta_y}(\mathbf{z}_i) \rightarrow \hat{y}_i$  to denote the adapter/classifier component of $f_\theta$ that maps the embedding $\mathbf{z}_i$ of a node $v_i$ to the predicted class $\hat{y}_i$. In addition, the self-supervised pretext task can be formulated to utilize the same feature extractor $f_{\theta_z}$ and an additional adapter $f_{\theta_s}$. Thus, the overall objective can be defined as follows, 
\begin{equation}
\min _{\theta,\theta^{\prime}} \mathcal{L}_{task}\left(\theta, \mathbf{A}, \mathbf{X}, \mathcal{D}_{L}\right) + \lambda\mathcal{L}_{self}\left(\theta^{\prime}, \mathbf{A}, \mathbf{X}, \mathcal{D}_{U}\right), 
\end{equation} 
where $\theta^{\prime} =\{\theta_z,\theta_s\}$ and $\lambda$ is the hyperparameter to control the contribution of self-supervision.

\begin{figure}[t]
    \centering
    \includegraphics[width=0.9\linewidth]{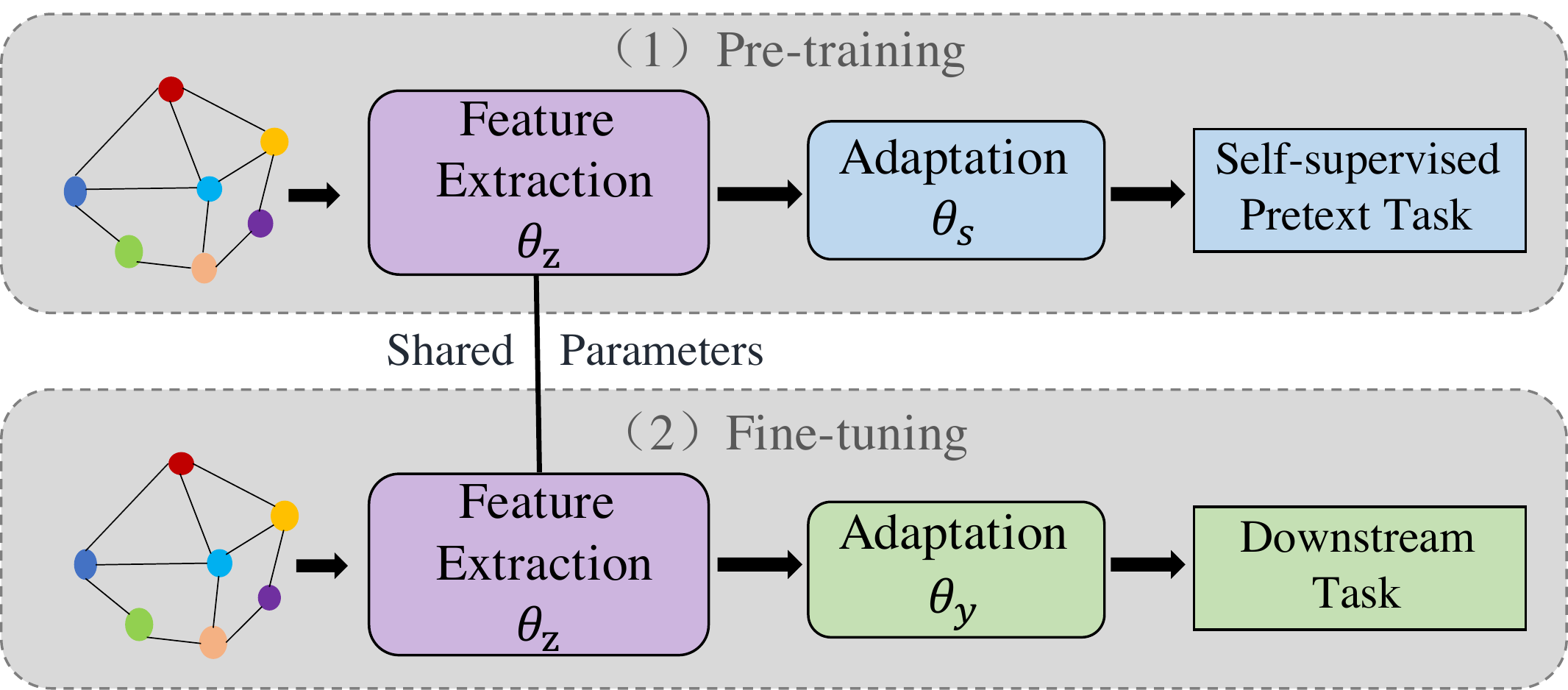}
    \caption{An overview of SSL in GNNs using two-stage training.}
    \label{fig:two-stage_training}
\end{figure} 

\subsection{Two-stage Training}
\label{sec:two-stage}
Two common strategies to utilize features learned via self-supervision in computer vision include applying the self-supervised model as an initialization for fine-tuning~\cite{zhai2019s4l, noroozi2018boosting} and  training a linear classifier over the learned features~\cite{kolesnikov2019revisiting,doersch2015unsupervised}. These strategies motivate us a two-stage training method to integrate SSL into GNNs. This method consists of the following two stages: 1) Pre-training on the self-supervised pretext task; and 2) Fine-tuning on the downstream task. 

An overview of the two-stage training method is given in Figure~\ref{fig:two-stage_training}. Similar to the joint training, the self-supervised model consists of a feature extraction module $f_{\theta_z}(\mathcal{G})$ and an adaptation module $f_{\theta_s}(\mathcal{G}))$, which are optimized by itself independent of the downstream task. Then, after the self-supervised model is fully trained,  we begin to train the downstream task model. More specifically, the downstream task model also has an adaptation module $f_{\theta_y}(\mathcal{G})$, but its feature extraction module shares parameters with that of the self-supervised model $f_{\theta_z}(\mathcal{G})$. As seen in the figure, we first pre-train the self-supervised model for the pretext task, and then use the self-supervised model's feature extraction module as the initialization of that of the downstream task. After initializing $\theta_z$, we can either fix it or fine tune it for the downstream task.

\subsection{Empirical Study}
In this subsection, we conduct extensive experiments based on the basic pretext tasks to understand {\it what} SSL information works for GNNs, {\it which} strategies can better integrate SSL for GNNs, and further analyze {\it why} SSL is able to improve GNNs. Following the setting in GCN~\cite{kipf2016semi}, we conduct experiments on the public data splits of three widely used benchmark datasets: Cora, Citeseer, and Pubmed. The dataset statistics can be found in Table~\ref{tab:data}.  We used Adam optimizer with learning rate 0.01, $L_2$ regularization 5e-4, dropout rate 0.5, 128 hidden units across all self-supervised information and GCN, and top-K = bottom-K = 3. Then parameters tuned on validation accuracy are: $\lambda$ in range of $\{0, 0.001, 0.01, 0.1, 1, 5, 10, 50, 100, 500, 1000\}$, $m_e$ and $m_a$ in \{10\%, 20\%\} the size of $|V|$. 

\begin{table}[t]
\centering
\setlength{\tabcolsep}{2pt}
\renewcommand{\arraystretch}{0.95}
\caption{Dataset statistics.}
\begin{tabular}{lcccccc}
\toprule
\textbf{Dataset} & \textbf{Nodes} & \textbf{Edges} & \textbf{Classes} & \textbf{Features} &\textbf{Training/Validation/Test} \\
\midrule Cora &  2,708 & 5,429 & 7 & 1,433 & 140/500/1000\\
Citeseer &  3,327 & 4,732 & 6 & 3,703 & 120/500/1000 \\
Pubmed &  19,717 & 44,338 & 3 & 500 &  60/500/1000\\
\bottomrule
\end{tabular}
\label{tab:data}
\end{table}

{\bf Strategies for Two-stage Training.} For two-stage training, after initializing the feature extractor, we can either fix it or fine tune it for node classification. We studied various architectures for both pretrained and node classification models with the results demonstrated in Table~\ref{tab:arc-choice}. Note that in the table, (1) ``2GC+1Linear" denotes that we use {\it two} graph convolutional layers for feature extraction and {\it one} linear layer for the adaptation; (2) ``2GC" means {\it one} graph convolutional layers for feature extraction and {\it another} graph convolutional layer for the adaptation; (3) for the column of ``Finetune Strategy", ``Fix" and ``Tune all" correspond to the aforementioned two strategies and we also report the performance of node classification without pretraining from SSL as the third strategy; and (4) all the experiments are conducted with the \textit{PairwiseDistance} task. In most cases, the strategy of ``Tune all" achieves the best performance. Thus, we choose this strategy when using two-stage training. We also note that the configuration of one graph convolutional layer for feature extraction, one graph convolutional layer for the adaptation of node classification and one linear layer for the adaptation of pretext task works very well for all three strategies. Therefore, we select this configuration for the remaining experiments unless stated otherwise.

\begin{table}[t]
\centering
\renewcommand{\arraystretch}{0.9}
\caption{ Two-stage training strategies on Cora.}
\begin{tabular}{@{}cccc@{}}
\toprule
\begin{tabular}{@{}c@{}} \textbf{SSL}\\ \textbf{Pretrained} \\ \textbf{Model}\end{tabular} & \begin{tabular}{@{}c@{}} \textbf{Node} \\ \textbf{Classification} \\ \textbf{Model}\end{tabular} & \textbf{\begin{tabular}[c]{@{}l@{}}Finetune Strategy\\      \end{tabular}} & \textbf{\begin{tabular}[c]{@{}c@{}}Test Accuracy\\  (\%)    \end{tabular}} \\ \midrule
2GC+1Linear                     & 2GC+1Linear                       & Fix                                                                                                        & 73.53            \\ 
2GC+1Linear                     & 2GC+1Linear                       & Tune all                                                                                                          & 80.55            \\ 
-                               & 2GC+1Linear                       & -                                                                                                                 & 78.63            \\ \midrule
2GC+1Linear                     & 3GC                           & Fix                                                                                                        & 74.69            \\     
2GC+1Linear                     & 3GC                             & Tune all                                                                                                          & 82.49   \\ 
-                               & 3GC                               & -                                                                                                                 & 80.88            \\ \midrule
1GC+1Linear                     & 1GC+1Linear                       & Fix                                                                                                        & 80.75            \\ 
1GC+1Linear                     & 1GC+1Linear                       & Tune all                                                                                                          & 79.79            \\ 
-                               & 1GC+1Linear                       & -                                                                                                                 & 78.75            \\ \midrule
1GC+1Linear                     & 2GC                               & Fix                                                                                                        & 81.04            \\ 
1GC+1Linear                     & 2GC                               & Tune all                                                                                                          & 82.39   \\ 
-                               & 2GC                               & -                                                                                                                 & 81.32            \\ \bottomrule
\end{tabular}
\label{tab:arc-choice}
\end{table}

{\bf SSL for GNNs.} Following the aforementioned experimental settings, we evaluate six basic pretext tasks in Section~\ref{sec:sslinfo} with joint and two-stage training strategies and the results are shown in Table~\ref{tab:analysis-performance}. 

{\it Joint Training vs. Two-stage Training.} We observe that although the two-stage training is able to improve the vanilla GCN model, the joint training outperforms the two-stage training in most settings. This observation from the graph domain is consistent with that from image self-supervised semi-supervised learning~\cite{zhai2019s4l}. In addition, the joint training strategy is less complicated as compared to the fine-tuning strategy. More specifically, joint training only requires the tuning of a single hyperparameter $\lambda$ as compared to significant efforts for the two-stage training due to the high sensitivity in the two-stage training as shown in Table~\ref{tab:arc-choice}. Thus, our empirical analysis suggests that joint training is a better strategy to integrate SSL with GNNs than the two-stage training.

{\it What SSL Works for GNNs.} From Table~\ref{tab:analysis-performance}, we can first observe that the best performance is always achieved by one including an SSL pretext task. In other words, our empirical analysis clearly shows that utilizing self-supervised information in graph neural networks is a promising direction for further improving the performance of deep learning on graph-structured data. Furthermore, we observe a wide range of utility for the various self-supervised pretext tasks for improving node classification. First, we notice that across all datasets, the best performing method is a pretext task developed from global structure information. Another thing to note is that we determine the quality of \textit{AttributeMask} in comparison to GCN-PCA, since they both first utilize PCA as a preprocessing step to reduce the dimension of the node features/attributes. Then, under further analysis on the results in Table~\ref{tab:analysis-performance}, we find that the pretext tasks of \textit{NodeProperty}, \textit{EdgeMask} and \textit{AttributeMask} cannot boost the original GCN since the performance difference is always smaller than $0.3\%$. By contrast, global self-supervision including \textit{PairwiseDistance}, \textit{Distance2Cluster}, and \textit{PairwiseAttrSim} successfully improves the performance (e.g., over $2\%$ improvement on the Cora dataset). Thus, self-supervised information from both the structure and attributes have potentials; while for the structure information, the global pretext tasks are likely to provide much more significant improvements compared to the local ones.

\begin{table}[t]
\setlength{\tabcolsep}{2pt}
\centering
\caption{Performance evaluation of using SSL for GNNs.}
\label{tab:analysis-performance}
\begin{tabular}{c|ccc|ccc}
\toprule
\multirow{2}{1cm}{\textbf{Model}} & \multicolumn{3}{c}{\textbf{\begin{tabular}[c]{@{}c@{}}Joint Training \end{tabular}}} & \multicolumn{3}{|c}{\textbf{\begin{tabular}[c]{@{}c@{}}Two-stage Training \end{tabular}}} \\ \cmidrule(r){2-4}  \cmidrule(r){5-7}
 & \textbf{Cora} & \textbf{Citeseer} & \textbf{Pubmed} & \textbf{Cora} & \textbf{Citeseer} & \textbf{Pubmed} \\ \hline
\textbf{GCN} & 81.32 & {71.53} & 79.28 & 81.32 & {71.53} & 79.28 \\ 
\textbf{GCN-DroppedGraph} & 81.03 & {71.29} & 79.28 & 81.03 & {71.29} & 79.26 \\ 
\textbf{GCN-PCA} & 81.74 & {70.38} & 78.83 & 81.74 & {70.38} & 78.83 \\ \hline
\textbf{NodeProperty} & 81.94 & 71.60 & 79.44 & 81.59 & 71.69 & 79.24 \\ 
\textbf{EdgeMask} & 81.69 & 71.51 & 78.90 & 81.44 & 71.57 & 79.33 \\ \hline
\textbf{PairwiseNodeDistance} & 83.11 & {71.90} & {80.05} & {82.39} & {72.02} & {79.57} \\ 
\textbf{Distance2Cluster} & {83.55} & 71.44 & 79.88 & 81.80 & 71.55 & 79.51 \\ \hline
\textbf{AttributeMask} & 81.47 & 70.57 & 78.88 & 81.31 & 70.40 & 78.72 \\ 
\textbf{PairwiseAttrSim} & 83.05 & 71.67 & 79.45 & 81.57& 71.74 & 79.42 \\ 
\bottomrule
\end{tabular}
\end{table}

\begin{figure}[t]
\centering
\renewcommand{\arraystretch}{0.95}
\vspace{-2.5ex}
\includegraphics[width=0.7\linewidth]{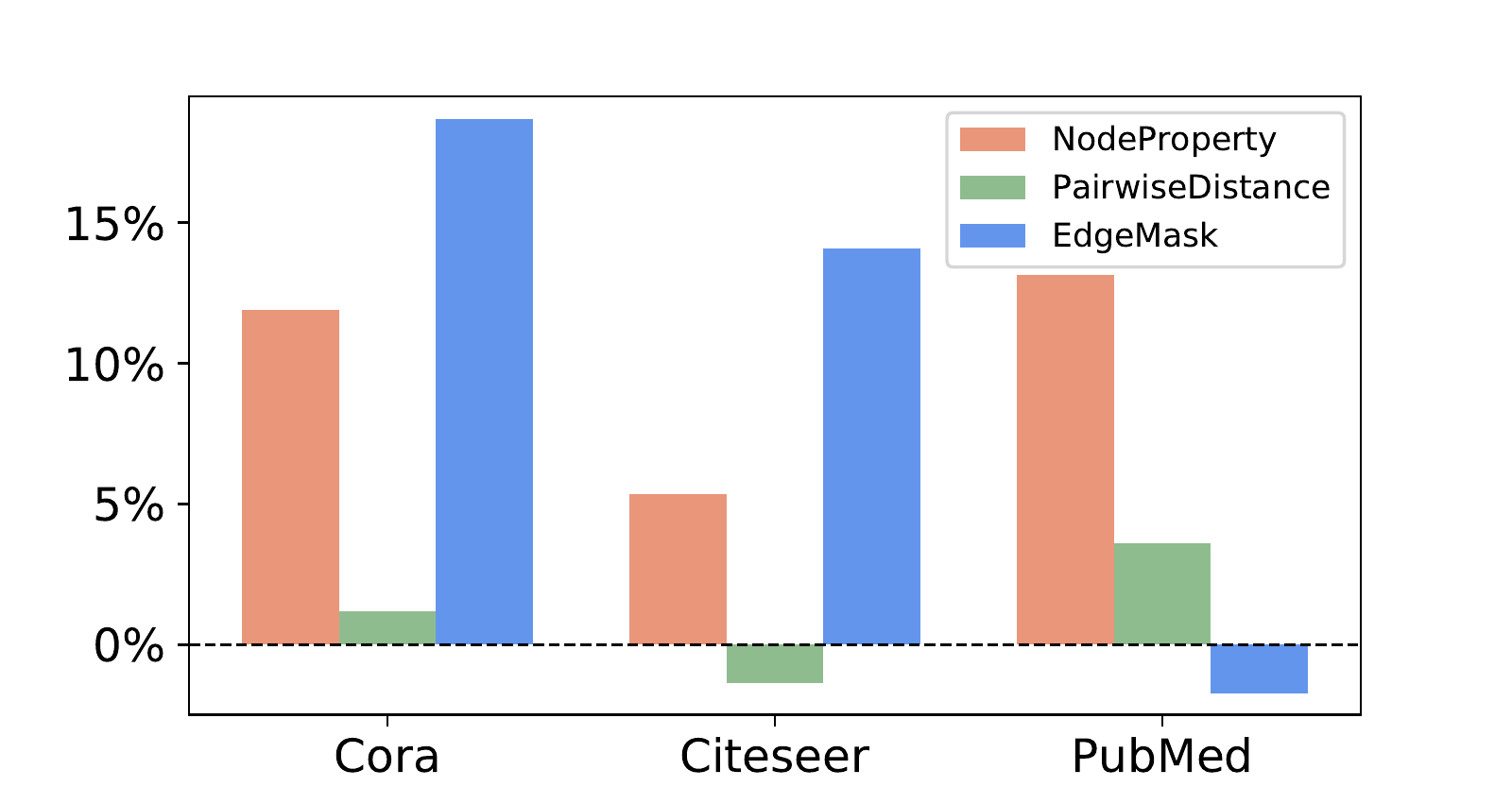}
\vspace{-1ex}
\caption{Classification accuracy difference on three SSL pretext tasks between GCN node representations and original node features. Positive value indicates the classifier trained with GCN node representations can achieve higher accuracy than that with original features, and vice versa.}
\label{fig:diff}
\vskip -1.75ex
\end{figure}

{\it Why SSL Works for GNNs.} We have found that using some self-supervision like global structure and pairwise attribute information can perform well while others not. Thus, the natural question is why they work or why they are unable to improve. As we mentioned before, GCN for node classification is naturally semi-supervised that has explored the unlabeled nodes. Therefore, one possible reason why some self-supervision cannot help could be that GCN can already learn such information itself. If this is the case, then training on additional self-supervised pretext task could perhaps not further boost the performance. To verify this assumption, we train logistic regression classifiers on the original node features and nodes representations from GCN (without self-supervised learning) to predict on the pretext task. The intuition is if GCN can learn one type of self-supervision, the nodes representations from GCN should have preserved such information; thus they should perform better on the corresponding pretext task than the original node features. The performance difference is shown in Fig~\ref{fig:diff}. We choose three representative pretext tasks \textit{EdgeMask}, \textit{NodeProperty} and \textit{PairwiseDistance} to illustrate since similar patterns can be observed in other cases. From the figure we can observe that GCN node representation consistently outperforms original features for \textit{NodeProperty} task by a large margin, indicating that GCN can learn such local structure information. Hence, \textit{NodeProperty} cannot bring in further improvement for its corresponding self-supervised GCN. Similar observations can be made on \textit{EdgeMask} task for Cora and Citeseer. The reason why the performance difference for \textit{EdgeMask} on Pubmed is small could be that the original features of Pubmed are very representative for local connectivity, since original features can achieve over 80\% accuracy on the \textit{EdgeMask} task. On the contrary, the performance difference on \textit{PariwiseDisance} task is rather small across three datasets. This observation suggests that GCN is unable to naturally learn the global structure information and employing pairwise node distance prediction as the SSL task can help boost its performance for the downstream task.

\textbf{Insights on SSL for GNNs.} One of the most fundamental properties in graphs is the study of graph similarity, which is to describe how similar two nodes are in a graph. There are two most popular approaches to measure this similarity including structural equivalence and regular equivalence~\cite{newman2018networks}. More specifically, one way of defining similarity in regards to structural equivalence in a graph is that two nodes are structurally equivalent if their local neighborhoods significantly overlap. In comparison, regular equivalent nodes are those that, while not necessarily having the same neighbors, have neighbors who themselves are similar~\cite{borgatti1992notions}. Now, given these definitions, we discuss and characterize to what extent GCN is able to naturally maintain these types of similarity when mapping from the input space to the embedding space, and then how these are correlated to the observations we have made with different self-supervised pretext tasks. 

First, given that GCN works by aggregating features from a node's local neighborhood, if two nodes have a significant overlap in their neighbors (as defined in structural equivalence), then it would be expected that their embeddings should somehow maintain this notion of similarity. Second, we let neighbor similarity in regular equivalence be defined in terms of their attribute similarity, then even if two nodes do not share the same neighbors, if their neighbors are similar, this will result in the learned embeddings of the two nodes maintaining this notion of regular attribute equivalence. 

Furthermore, we can observe our proposed self-supervised pretext tasks built on structure information (e.g., {\it PairwiseNodeDistance}) can be described in helping to maintain this notion of structural equivalence in the embeddings. For example, if the embeddings are encouraged to encode how similar two nodes are in terms of their distance (as done in {\it PairwiseNodeDistance}), then this is related to maintaining structural similarity as many node similarity measures are defined based on the idea of path length between two nodes. For the self-supervised pretext tasks based on attribute information, if we define the concept of attribute equivalence being two vertices that share many of the same attributes/features (as structural share many neighbors). Then, given this definition of attribute equivalence, we can observe that indeed self-supervised pretext tasks based on attribute information such as {\it PairwiseAttrSim} are actually looking to explicitly maintain this notion of similarity from the input to the embedding space. 

With these insights and our empirical observations, we are able to design more pretext tasks. In this work, we aim to try a new direction beyond the structure and attribute information where we want to take into consideration the specific downstream task. In particular, we want to extend the concept of regular equivalence being defined with similarity on the level of structure or attributes to instead the level of task by introducing {\it regular task equivalence} where node similarity is now defined specific to the task. Since our downstream task is node classification, beyond the structure and attribute information (i.e., ${\bf A}$ and ${\bf X}$, respectively), we additionally have the label information for some of the nodes (i.e., those in $\mathcal{D}_L$). Thus, in the next section, we will discuss advanced self-supervised methods that are built with the intuition of adapting the notion of regular equivalence beyond having neighbors with similar attributes, to instead having neighbors with similar node labels (or regular task equivalence). More specifically, the general idea is that if every node constructs a pretext vector based on information in regards to the labels from their neighborhood, then two nodes having similar (or dissimilar) vectors, we would encourage to be similar (or dissimilar) in the embedding space. {\it The significance of this new attempt is that if the concept of regular task equivalence can work, it will open new doors to design more advanced pretext tasks based on the concept of equivalence from each individual resource or their combinations. }
\section{Advanced Pretext Tasks on Graphs}\label{sec:ssladvanced}

Based on our analysis from Section~\ref{sec:sslstudy}, constructing a self-supervised pretext task that helps the downstream task from only structure or attribute information is not always able to find additional improvement, since such information could have already been partially maintained by GCN. Thus, given that there can be different downstream tasks and associated information (e.g., a small set of labeled nodes), we can directly exploit the task specific self-supervised information referred as {\it SelfTask} in this work. In particular, we will develop various pretext tasks of {\it SelfTask} that extend the idea to maintain regular task equivalence defined over the set of class labels.

\subsection{SelfTask: Distance2Labeled} 

We first investigate modifying one of the best performing global structure pretext tasks, {\it Distance2Cluster} to take into consideration information from the labeled nodes. To incorporate label information with graph structure information, we propose to predict the distance vector from each node to the labeled nodes (i.e., $\mathcal{V}_L$) as the SSL task. For class $c_j\in\{1,...,K\}$ and unlabeled node $v_i\in \mathcal{V}_U$, we calculate the average, minimum and maximum shortest path length from $v_i$ to all labeled nodes in class $c_i$. Thus, the distance vector for $v_i$ can be denoted as ${\bf d}_i\in\mathbb{R}^{3K}$. Formally, the SSL objective can be formulated as a multiple regression problem as follows,
    \begin{equation}
    \mathcal{L}_{self}( {\theta}^{\prime}, {\bf A}, {\bf X}, \mathcal{D}_U) = \frac{1}{\vert{\mathcal{D}_U}\vert}\sum_{ v_i \in \mathcal{D}_U} \|f_{\theta^{\prime}}(\mathcal{G})_{v_i} - {\bf d}_{i}\|^2. 
    \end{equation}   
    This formulation can be seen as a way of strengthening the global structure pretext, but mostly focused on leveraging the task specific information of labeled nodes. 
    
\subsection{SelfTask: ContextLabel} 

In Section~\ref{sec:sslstudy}, we analyzed the types of similarity that GNNs are naturally positioned to maintain and how our basic pretext tasks are able to further improve them, namely those based on structure and attributes. This led to our definition of regular task equivalence, which if maintained would imply that nodes who have neighbors with similar labels should themselves be similar in the embedding space. However, in the node classification task this would require us to have labels for our neighbors so that we can harness this new concept of similarity in self-supervised pretext tasks. However, labels often are sparse for the task of node classification. Therefore, we propose to use a similarity based function. It can utilize structure, attributes, and the current labeled nodes to construct a neighbor label distribution context vector ${\bf \bar{y}}_i$ for each node as follows,
    \begin{align}
    f_s({\bf A}, {\bf X}, \mathcal{D}_L, \mathcal{V}_U) \rightarrow  \{ {\bf \bar{y}}_i | v_i \in \mathcal{V}_U\},
    \end{align}
More specifically, we define the context of node $v_i$ as all nodes within its $k$-hop neighbors where the $c$-th element of the label distribution vector can be defined as,
    \begin{equation}
        \label{eq:contextlabel}
    {\bf \bar{y}}_{ic} = \frac{|\mathcal{N}_{\mathcal{V}_L}(v_i,c)| + |\mathcal{N}_{\mathcal{V}_U}(v_i,c)|}{|\mathcal{N}_{\mathcal{V}_L}(v_i)| + |\mathcal{N}_{\mathcal{V}_U}(v_i)|}, c={1\ldots{}K},
    \end{equation}
    where $\mathcal{N}_{\mathcal{V}_U}(v_i)$ denotes the neighborhood set from $\mathcal{V}_U$ of node $v_i$,  $\mathcal{N}_{\mathcal{V}_U}(v_i,c)$ then denotes only those in the neighborhood set having been assigned class $c$
    (with similar definitions for $\mathcal{V}_L$ neighborhood sets), and $\mathcal{D}_U=(\mathcal{V}_U, \{ {\bf \bar{y}}_i | v_i \in \mathcal{V}_U\} )$. 
    Furthermore, the objective function for this pretext task based on the concept of regular task equivalence can then be formulated as, 
    \begin{equation}
    \label{eq:contextobj}
    \mathcal{L}_{self}( {\theta}^{\prime}, {\bf A}, {\bf X}, \mathcal{D}_U) = \frac{1}{\vert{\mathcal{D}_U}\vert}\sum_{ v_i \in \mathcal{D}_U} \|f_{\theta^{\prime}}(\mathcal{G})_{v_i} - {\bf \bar{y}}_{i}\|^2. 
    \end{equation}   
 Several methods can be selected for extending the label information to all unlabeled nodes for $f_s$. One way is to use a method based on structure equivalence where we elect to use Label Propagation (LP)~\cite{zhu2003semi} since it only uses {\bf A} (although others like shortest path could be extended here as used in {\it Distance2Label}). Another way is using both structure equivalence and attribute equivalence where we use the Iterative Classification Algorithm (ICA)~\cite{sen2008collective} that utilizes both {\bf A} and {\bf X}.  The neighbor label distribution context vectors $\{{\bf \bar{y}}_i | v_i \in \mathcal{V}_U\} $ could be noisy due to the inclusion of weak labels 
 produced by $f_s$ (e.g., when using LP or ICA). Next we introduce two methods to improve {\it ContextLabel}.

\subsection{SelfTask: EnsembleLabel}

There are various ways to define the similarity based function $f_s$ such as LP and ICA. Thus, one possible way to improve {\it ContextLabel} is to ensemble various functions $f_s$. If we let the class probabilities for a node $v_i$ to be $\sigma_{LP}(v_i)$ and $\sigma_{ICA}(v_i)$, respectively when using LP and ICA inside $f_s$, then we can combine them to select $\bar{y}_i$ as, 
\begin{equation}
\bar{y}_i=\operatorname{argmax}_{c} \sigma_{LP}(v_i) + \sigma_{ICA}(v_i), \quad c=1 \ldots K
\end{equation}
We can use the ensembled $\bar{y}$ for constructing context label distribution like Eq.~(\ref{eq:contextlabel}) and following the pretext objective defined in Eq.~(\ref{eq:contextobj}).

\subsection{SelfTask: CorrectedLabel} 

We design {\it CorrectedLabel} as an alternative pretext task to enhance {\it ContextLabel} by iteratively improving the context vectors.  More specifically, we take the approach of iteratively training the GNN and correcting the labels ${\bf \hat{y}}_{i}$ similar to the iterative training in \cite{han2019deep-noisy} using training and correction phases. In the training phase, we use the corrected label to build the corrected context label distribution vector for unlabeled nodes similar to Eq.~(\ref{eq:contextlabel}).
We use $\mathcal{\hat{D}}_U=(\mathcal{V}_U,\{ {\bf \hat{y}}_i | v_i \in \mathcal{V}_U\})$ to denote the unlabeled data samples with its corrected context label distribution in addition to $\mathcal{D}_U=(\mathcal{V}_U,\{ {\bf \bar{y}}_i | v_i \in \mathcal{V}_U\})$ for the SSL task. 
Then GNN $f_\theta$ is trained on both the original (e.g., ${\bf \bar{y}}_{i}$) and corrected (e.g., ${\bf \hat{y}}_{i}$) context distributions where the loss can be formulated as,

\vspace{-1ex}
\begin{footnotesize}
\begin{align}
\mathcal{L}_{self}\left(\theta^{\prime}, \mathbf{A}, \mathbf{X}, \mathcal{D}_{U}, \hat{\mathcal{D}}_{U}\right) &= \frac{1}{\vert{\mathcal{D}_U}\vert}\sum_{ v_i \in \mathcal{D}_U} \|f_{\theta^{\prime}}(\mathcal{G})_{v_i} - {\bf \bar{y}}_{i}\|^2  \\
&+ \alpha \Bigg( \frac{1}{\vert{\mathcal{D}_U}\vert}\sum_{ v_i \in \mathcal{D}_U} \|f_{\theta^{\prime}}(\mathcal{G})_{v_i} - {\bf \hat{y}}_{i}\|^2 \Bigg), \nonumber 
 \end{align} 
 \end{footnotesize}
 
\noindent where the first and second terms are to fit the original and corrected context distributions respectively, and $\alpha$ controls the contribution from the corrected context distributions.  

In the label correction phase, we employ the trained GCN to select $p$ class prototypes ${\bf Z_c}=\{{\bf z}_{c1},\dots,{\bf z}_{cp}\}$ (represented as deep features) for each category $c$, which are used to generate the corrected label. More specifically, we first randomly sample $m$ nodes in the same class to calculate their pair-wise similarity matrix ${\bf S}$ where ${\bf S}_{ij}$ is the cosine similarity between two nodes based on their embeddings. Then we define density $\rho_i$ for each node as,
\begin{equation}
\rho_{i}=\sum_{j=1}^{m} \operatorname{sign}\left({\bf S}_{i j}-S_{c}\right),
\end{equation}
where $S_c$ is a constant value (which we selected as the value ranked in top 40\% in ${\bf S}$ as suggested in~\cite{han2019deep-noisy}) and $\operatorname{sign}(x) = 1, 0,$ or $-1$ if $x >0, =0$, or $<0$, respectively. According to the formulation, a smaller $\rho$ indicates that the node is less similar to other nodes in the same class. The nodes with inconsistent labels are usually isolated from others while nodes with correct labels should be close to each other. Hence, we select the nodes with top-$p$ largest $\rho$ values as the class prototypes. Then we calculate the corrected label $\hat{y}_i\in \{1,\dots, K\}$ for node $v_i \in \hat{\mathcal{V}}_U$ as, 
\begin{equation}
\hat{y}_i=\operatorname{argmax}_{c} \frac{1}{p} \sum_{l=1}^{p} \cos \left(f_{\theta^{\prime}}(\mathcal{G})_{v_i}, {\bf z}_{cl}\right), c=1 \ldots K,
\end{equation}
where $\cos(\cdot,\cdot)$ is used to denote the cosine similarity between two samples. In other words, we use the average similarity between $v_i$ and $p$ prototypes to represent the similarity between $v_i$ and the corresponding class, and then assign the class $c$ having the largest similarity to $v_i$. By iterating the two phases, the GNN can gradually learn corrected labels (e.g., $\hat{y}_i$).

\section{Experiments}\label{sec:experiments}
In this section, we evaluate the effectiveness of the proposed SelfTask pretext tasks presented in Section~\ref{sec:ssladvanced}. Before presenting our experimental results and observations, we first introduce the experimental settings.

\subsection{Experimental Settings}

To validate the proposed approaches, we conduct experiments on four benchmark datasets, including Cora, Citeseer and Pubmed~\cite{kipf2016semi} shown in Table~\ref{tab:data}, and Reddit~\cite{hamilton2017inductive}. More specifically, Reddit has 232,965 nodes, 57,307,946 edges, 210 classes, 5,414 node features and training/validation/test node split as 152,410/23,699/55,334, respectively. We note that all experiments are performed in the transductive setting. 

We adopt 2-layer GCN as the backbone for node classification model, with hidden units of 128, $L_2$ regularization $5e{-4}$, dropout rate $0.5$ and learning rate $0.01$. For the SSL loss, we take out the hidden representations from the first layer of GCN and feed them through a linear layer to solve SSL pretext task. We utilize the strategy of jointing training to integrate SSL with GCNs. The weighting parameter $\lambda$ for joint training is searched from $\{1, 5, 10, 50, 100, 500\}$. The parameter $\alpha$ of {\it CorrectedLabel} is searched from $\{0.5, 0.8, 1, 1.2, 1.5\}$. For \textit{ContextLabel}, \textit{EnsembleLabel}, and \textit{CorrectedLabel}, the neighborhood range $k$ is set to 2 for Cora, Citeseer and Pubmed, and 1 for Reddit. All the experiments are repeated 10 times and we report the average accuracy with standard deviation. The hyper-parameters of all the models are tuned based on the loss and accuracy on the validation set. In addition to the vanilla 2-layer GCN~\cite{kipf2016semi},  we also include two recent SSL methods on graph neural networks as baselines -- (1) \textbf{Self-Training}~\cite{li2018deeper-selftrain}: it first trains a graph neural network and adds the most confident predictions of unlabeled data to the label set as pseudo-labels for later training; and (2) \textbf{M3S}~\cite{sun2019m3s}: it repeatedly assigns pseudo-labels and trains on augmented labeled set for $K$ times where it employs DeepCluster~\cite{caron2018deepcluster} and Self-Training to perform self-checking based on the generated pseudo-labels.

\begin{table*}[t]
\renewcommand{\arraystretch}{0.95}
\setlength{\tabcolsep}{4.5pt}
\centering
\caption{Node classification performance accuracy (\%) of integrating SSL into GNNs.}
\label{tab:performance}
\begin{threeparttable}
\begin{tabular}{ccccc}
\toprule
\textbf{Model} & \textbf{Cora} & \textbf{Citeseer} & \textbf{Pubmed} & \textbf{Reddit} \\ \midrule
\textbf{GCN} & $81.32 \pm 0.33$ & $71.53 \pm 0.27$ & $79.28 \pm 0.44$ & $94.99 \pm 0.04$ \\ 
\textbf{Self-Training} & $81.29 \pm 0.33$ & $72.21 \pm 0.53$ & $79.35 \pm 0.62$ & $95.00 \pm 0.03$ \\ 
\textbf{M3S} & $81.59 \pm 0.41$ & $71.72 \pm 0.36$ & $79.50 \pm 0.76$ & $95.18\pm0.02$ \\ 
\midrule
\textbf{SelfTask-Distance2Labeled}\footnote{} & $\mathbf{83.39} \pm 0.40$ & $71.64 \pm 0.28$ &	$79.51 \pm 0.32$ & - \\ 
\textbf{SelTask-ContextLabel-LP} & $82.20 \pm 0.29$ & $71.90 \pm 0.20$ & $80.07 \pm 0.18$ & $95.32 \pm 0.02$ \\ 
\textbf{SelfTask-ContextLabel-ICA} & $82.76 \pm 0.46$ & $72.59 \pm 0.45$ & $82.31 \pm 0.14$ & $95.21 \pm 0.08$ \\ \midrule 
\textbf{SelfTask-EnsembleLabel} & $82.85 \pm 0.51$ & $72.48 \pm 0.28$ & $81.31 \pm 0.16$ &	$95.29 \pm 0.02$ \\ 
\textbf{SelfTask-CorrectedLabel-LP} & $82.63 \pm 0.33$ & $72.30 \pm 0.53$ & $80.16 \pm 0.30$ & $\mathbf{95.33} \pm 0.02$ \\ 
\textbf{SelfTask-CorrectedLabel-ICA} & $83.28 \pm 0.36$ & $\mathbf{73.04} \pm 0.35$ & $\mathbf{82.60} \pm 0.32$ & $95.25 \pm 0.09$ \\ 
\bottomrule
\end{tabular}
\begin{tablenotes}
  \item[1] \textit{SelfTask-Distance2Labeled} is not scalable on the Reddit dataset where the labeled/unlabeled data is huge, since as defined it requires calculating the shortest path length distance from labeled data to unlabeled data.
\end{tablenotes}
\end{threeparttable}
  \vspace{-0.2in}
\end{table*}

\subsection{Performance Comparison}
The node classification performance is demonstrated in  Table~\ref{tab:performance}. We first note that most of pretext tasks of SelfTask outperform existing SSL methods, i.e., Self-Training and M3S. This observation not only demonstrates the effectiveness of SelfTask but also indicates that the deep insights from the preliminary analysis have tremendous potentials to inspire new pretext tasks on graphs.  We observe that the pretext tasks based on \textit{ContextLabel} consistently improve GCN across all datasets by a large margin. For instance,  \textit{SelfTask-ContextLabel-ICA} improves GCN by $1.4\%$, $1.1\%$ and $3.0\%$ on Cora, Citeseer and Pubmed datasets respectively, achieving the state-of-the-art performance. By contrast, most of the basic SSL tasks can only improve on one dataset or achieve small improvement, which demonstrates the importance of task specific information in constructing stronger pretext tasks. Moreover, label correction consistently boosts the performance of \textit{SelfTask} on all datasets while label ensemble can only boost \textit{SelfTask-ContextLabel-LP} for most of the time. This observation  indicates that label correction can better extend label information to unlabeled nodes than ensemble. However, label correction is much less inefficient as it will optimize the process of correcting labels for unlabeled nodes. Hence, in practice, we need to balance the computational efficiency and predictive accuracy when choosing the best strategy.

\subsection{Fewer Labeled Samples}

\begin{figure}[t]
\centering
\includegraphics[width=0.8\linewidth]{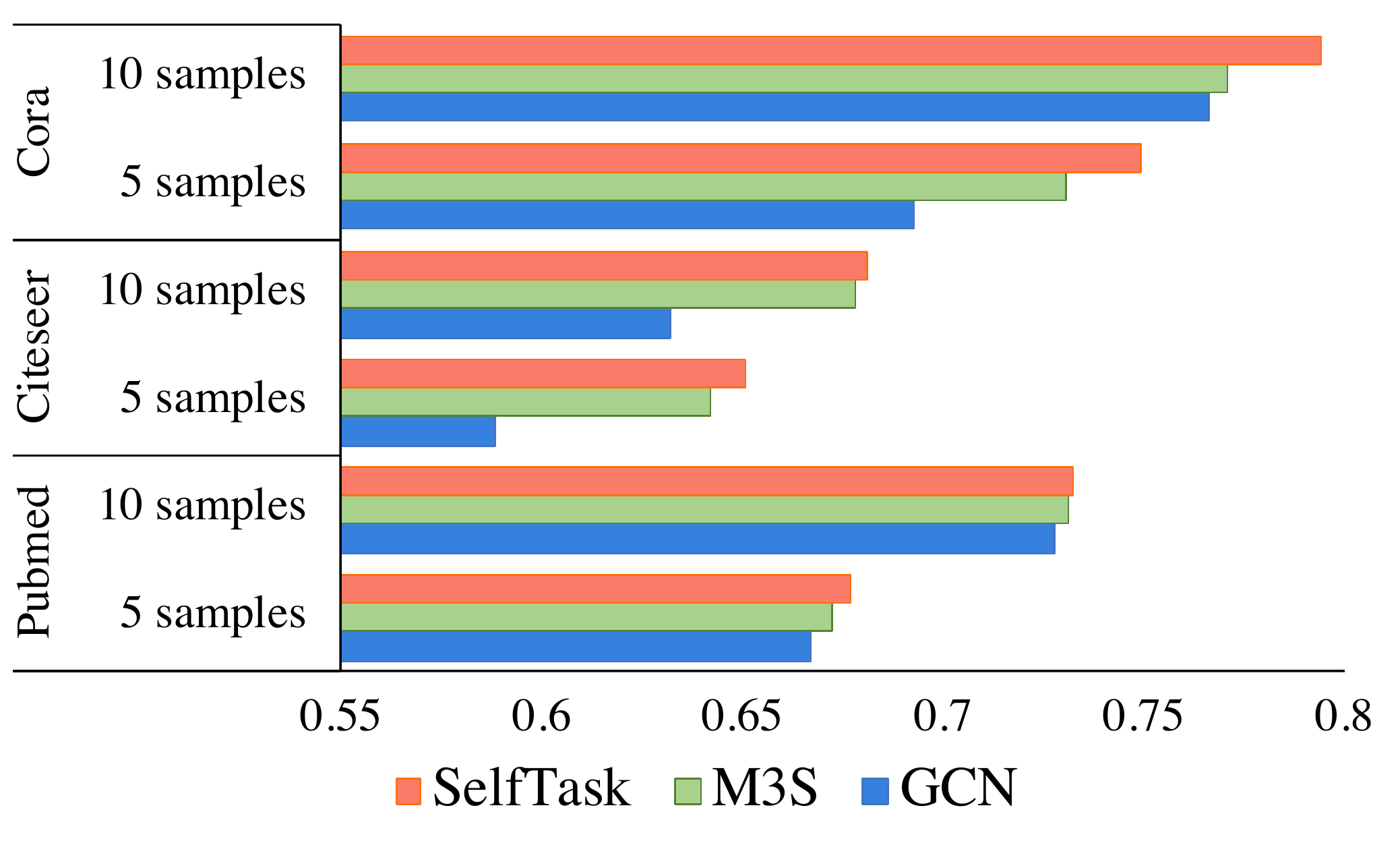}
\vskip -0.5ex
\caption{Accuracy performance with limited labeled nodes.}
\label{fig:few-label}
\end{figure}

The proposed SelfTask pretext tasks depends on the label information. In this subsection, we examine if SelfTask can still work when having a very limited number labeled samples. We randomly sampled 5 or 10 nodes per class for training and the same number of nodes for validation. All remaining labeled nodes are used testing. We repeated this process for 10 times and compare our best model, denoted as \textit{SelfTask} with GCN and M3S. Since the performance of Self-Training is always worse than that of M3S, we do not include its performance. The results are shown in Figure~\ref{fig:few-label}. As we can see from the figure, the performance of GCN drops rapidly with the decrease of labeled samples. However, \textit{SelfTask} achieves even greater improvement when the labeled samples are fewer and consistently outperforms the state-of-the-art baselines. Especially under the setting of 5 samples per class on Citeseer dataset, our proposed model improves GCN by a large margin of $7.2\%$. These observations suggest that SelfTask can be applied to the scenarios when labels are very sparse.

\subsection{Parameter Analysis}
\begin{figure}[tb]%
     \centering
     \subfloat[$\lambda$]{{\includegraphics[width=0.48\linewidth]{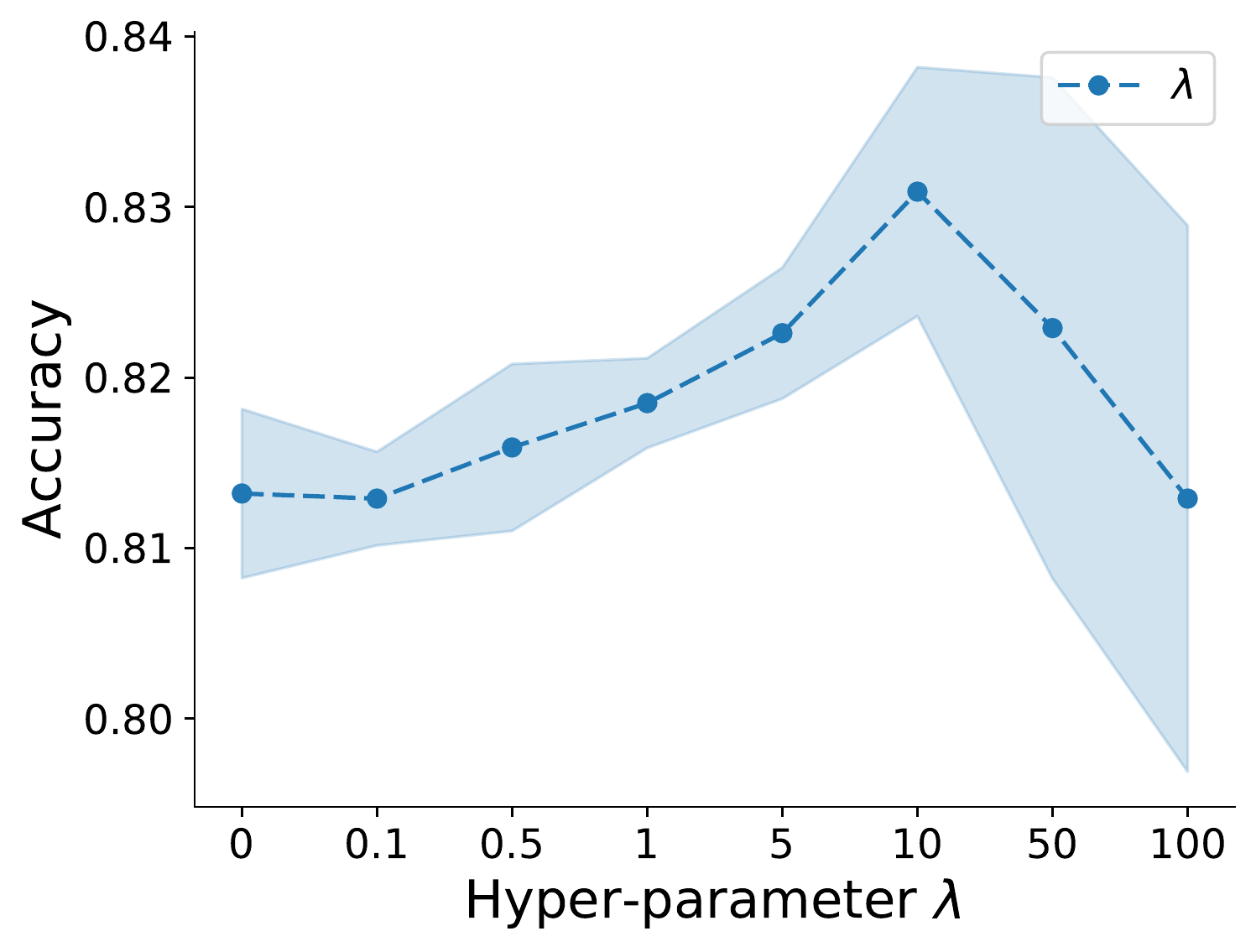} }\label{fig:param-lambda}}%
     \subfloat[$\alpha$]{{\includegraphics[width=0.48\linewidth]{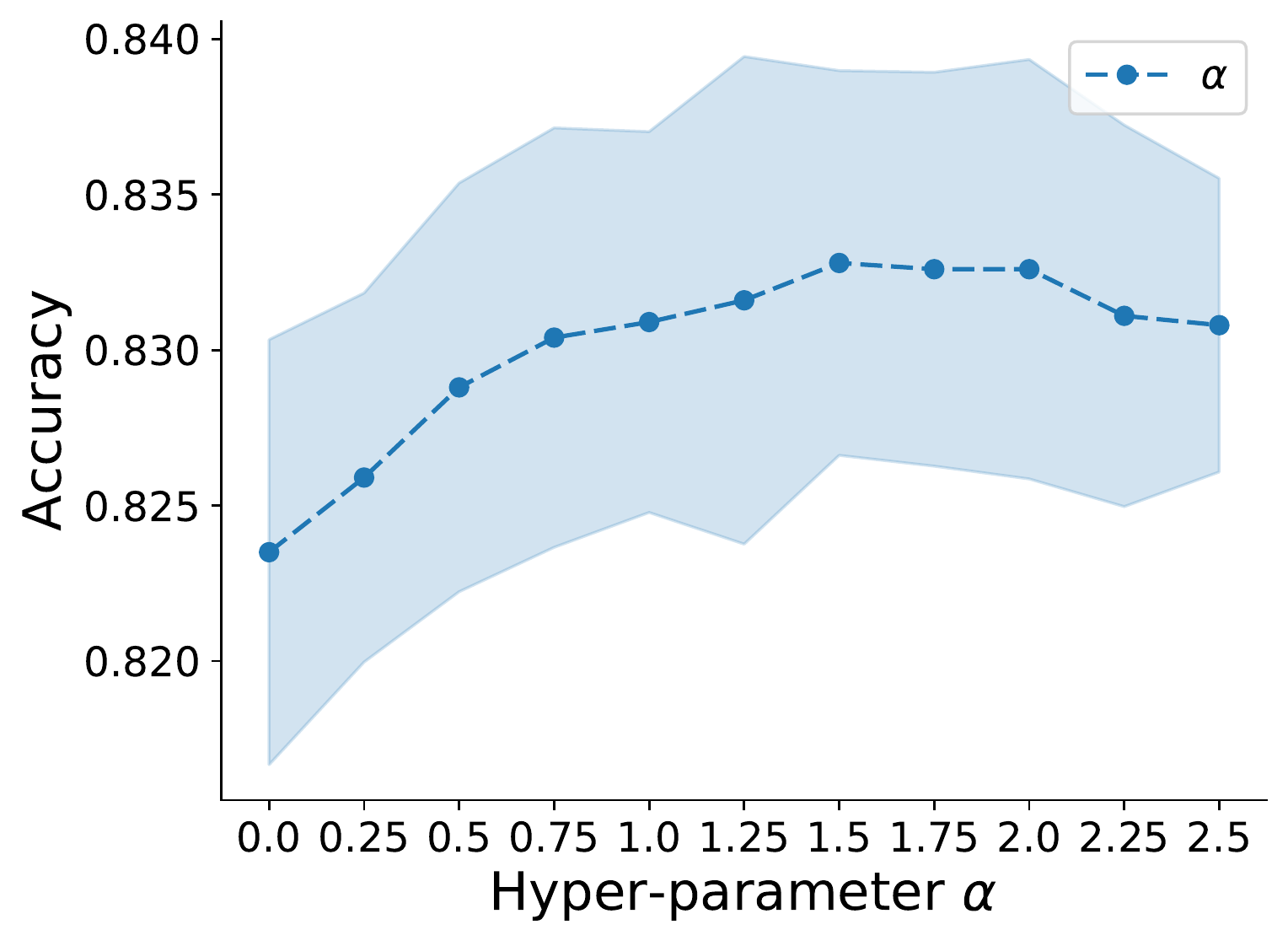} }\label{fig:param-alpha}}%
    \qquad
    \caption{Parameter analysis on Cora.} 
    \label{fig:param}
    \vskip -0.5em
\label{fig:ablation}%
\end{figure}

In this subsection, we explore the sensitivity of hyper-parameters for the best model,  \textit{SelfTask-CorrectedLabel-ICA}. Here we alter the value of $\lambda$ and $\alpha$ to see the changes of the model in terms of test accuracy. More concretely, we vary $\lambda$ in the range of $\{0, 0.1, 0.5, 1, 5, 10, 50, 100\}$ and $\alpha$ from $0$ to $2.5$ with an interval of $0.25$. We only report the results on the Cora dataset since similar patterns are observed in other datasets. The accuracy change in terms of $\lambda$ is illustrated in Figure~\ref{fig:param-lambda}. We can see the performance of our model first increases with the increase of $\lambda$. This result supports that incorporating SSL can boost the performance of GNNs. However, when $\lambda$ is large, the performance reduces due to the overfitting on the SSL task. Figure~\ref{fig:param-alpha} shows the impact of $\alpha$. Employing label correction ($\alpha>0$) outperforms not using it ($\alpha=0$), which suggests the effectiveness of label correction. 

\section{Related Work}\label{sec:relatedwork}
In this section, we introduce the related work including self-supervised learning and graph neural networks.

\subsection{Self-supervised Learning}

SSL is a novel learning framework that generates additional supervised signals to train deep learning models through carefully designed pretext tasks. SSL has been proven to effectively alleviate the problem of lack of labeled training data~\cite{kolesnikov2019revisiting}. In the image domain, various self-supervised learning techniques have been developed for learning high-level image representations. Doersch et al.~\cite{doersch2015unsupervised} first proposed to predict the relative locations of image patches. Following this line of research, Noroozi et al.~\cite{noroozi2016unsupervised} designed a pretext task called Jigsaw Puzzle. More types of pretext tasks have also been investigated, such as image rotation~\cite{gidaris2018rotation}, image clustering~\cite{caron2018deepcluster}, image inpainting~\cite{pathak2016context}, image colorization~\cite{zhang2016colorful} and motion segmentation prediction~\cite{pathak2017learning}. In the domain of graphs, there are a few works incorporating SSL. Sun et al.~\cite{sun2019m3s} utilized the clustering assignments of node embeddings as guidance to update the graph neural networks. Peng et al.~\cite{peng2020self} proposed to use the global context of nodes as the supervisory signals to learn node embeddings.  

\subsection{Graph Neural Networks}

GNNs can be roughly categorized into spectral methods and spatial methods. Spectral methods were initially developed based on spectral theory~\cite{bruna2013spectral,defferrard2016convolutional,kipf2016semi}. 
Bruna et al.~\cite{bruna2013spectral} first extended the notion of convolution to non-grid structures.  Afterward, a simplified version of spectral GNNs called ChebNet~\cite{defferrard2016convolutional} was developed. Next, GCN is proposed by Kipf et al.~\cite{kipf2016semi}, where Chebnet is further simplified based on its first-order approximation. Later, Wu et al.~\cite{wu2019simplifying} proposed Simple Graph Convolution (SGC) to simplify GCN by removing nonlinearities and collapsing weight matrices. Spatial methods consider the topological structure of the graph, and aggregate the information of nodes according to local information~\cite{hamilton2017inductive,gat}. 
Hamilton et al.~\cite{hamilton2017inductive} proposed an inductive learning method called GraphSAGE for large-scale networks. Veličković et al.~\cite{gat} proposed graph attention network (GAT), which includes an attention mechanism to graph convolutions. Further, Rong et al.~\cite{rong2019truly} developed deep graph convolution network by applying DropEdge mechanism to randomly drop edges during training. For a thorough review, please refer to recent surveys~\cite{wu2020comprehensive,zhou2018graph}. 
\section{Conclusion}\label{sec:conclusion}

Applying self-supervised learning to GNNs is a cutting-edge research topic with great potential. To facilitate this line of research, we have carefully studied SSL in GNNs for the task of node classification. We first introduce various basic SSL pretext tasks for graphs and present detailed empirical study to understand when and why SSL works for GNNs and which strategy can better work with GNNs. Next, based on our insights, we propose a new direction \textit{SelfTask} to build advanced pretext tasks which further exploit task-specific self-supervised information. Extensive experiments on real-world datasets demonstrate that our advanced method achieves state-of-the-art performance. Future work can be done on exploring new pretext tasks and applying the proposed SSL strategies in pre-training graph neural networks.

\newpage

\medskip

\small

\bibliographystyle{unsrt}
\bibliography{main}

\end{document}